# Automated Classification of L/R Hand Movement EEG Signals using Advanced Feature Extraction and Machine Learning

Mohammad H. Alomari, Aya Samaha, and Khaled AlKamha
Applied Science University
Amman, Jordan

*Abstract*— In this paper, we propose an automated computer platform for the purpose of classifying Electroencephalography (EEG) signals associated with left and right hand movements using a hybrid system that uses advanced feature extraction techniques and machine learning algorithms. It is known that EEG represents the brain activity by the electrical voltage fluctuations along the scalp, and Brain-Computer Interface (BCI) is a device that enables the use of the brain's neural activity to communicate with others or to control machines, artificial limbs, or robots without direct physical movements. In our research work, we aspired to find the best feature extraction method that enables the differentiation between left and right executed fist movements through various classification algorithms. The EEG dataset used in this research was created and contributed to PhysioNet by the developers of the BCI2000 instrumentation system. Data was preprocessed using the EEGLAB MATLAB toolbox and artifacts removal was done using AAR. Data was epoched on the basis of Event-Related (De) Synchronization (ERD/ERS) and movement-related cortical potentials (MRCP) features. Mu/beta rhythms were isolated for the ERD/ERS analysis and delta rhythms were isolated for the MRCP analysis. The Independent Component Analysis (ICA) spatial filter was applied on related channels for noise reduction and isolation of both artifactually and neutrally generated EEG sources. The final feature vector included the ERD, ERS, and MRCP features in addition to the mean, power and energy of the activations of the resulting Independent Components (ICs) of the epoched feature datasets. The datasets were inputted into two machine-learning algorithms: Neural Networks (NNs) and Support Vector Machines (SVMs). Intensive experiments were carried out and optimum classification performances of 89.8 and 97.1 were obtained using NN and SVM, respectively. This research shows that this method of feature extraction holds some promise for the classification of various pairs of motor movements, which can be used in a BCI context to mentally control a computer or machine.

## I. INTRODUCTION

The importance of understanding brain waves is increasing with the ongoing growth in the Brain-Computer Interface (BCI) field, and as computerized systems are becoming one of the main tools for making people's lives easier, BCI or Brain-Machine Interface (BMI) has become an attractive field of research and applications, BCI is a device that enables the use of the brain's neural activity to communicate with others or to control machines, artificial limbs, or robots without direct physical movements [1-4].

The term "Electroencephalography" (EEG) is the process of measuring the brain's neural activity as electrical voltage fluctuations along the scalp that results from the current flows in brain's neurons [5]. In a typical EEG test, electrodes are fixed on the scalp to monitor and record the brain's electrical activity [6]. BCI measures EEG signals associated with the user's activity then applies different signal processing algorithms for the purpose of translating the recorded signals into control commands for different applications [7].

The most important application for BCI is helping disabled individuals by offering a new way of communication with the external environment [8]. Many BCI applications were described in [9] including controlling devices like video games and personal computers using thoughts translation. BCI is a highly interdisciplinary research topic that combines medicine, neurology, psychology, rehabilitation engineering, Human-Computer Interaction (HCI), signal processing and machine learning [10].

The strength of BCI applications lies in the way we translate the neural patterns extracted from EEG into machine commands. The improvement of the interpretation of these EEG signals has become the goal of many researchers; hence, our research work explores the possibility of multi-trial EEG classification between left and right hand movements in an offline manner, which will enormously smooth the path leading to online classification and reading of executed movements, leading us to what we can technically call "Reading Minds".

In this work, we introduce an automated computer system that uses advanced feature extraction techniques to identify some of the brain activity patterns, especially for the left and right hand movements. The system then uses machine learning algorithms to extract the knowledge embedded in the recorded patterns and provides the required decision rules for translating thoughts into commands (as seen in Fig. 1).

This article is organized as follows: a brief review of related research work is provided in Section II. In Section III, the dataset used in this study is described. The automated feature extraction process is described in Section IV. The generation of our training/testing datasets and the practical implementation and system evaluation are discussed in Section V. Conclusions and suggested future work are provided in Section VI.

## II. LITERATURE REVIEW

The idea of BCI was originally proposed by Jaques Vidal in [11] where he proved that signals recorded from brain activity could be used to effectively represent a user's intent. In [12], the authors recorded EEG signals for three subjects while imagining either right or left hand movement based on a visual cue stimulus. They were able to classify EEG signals into right and left hand movements using a neural network classifier with an accuracy of 80% and concluded that this accuracy did not improve with increasing number of sessions.

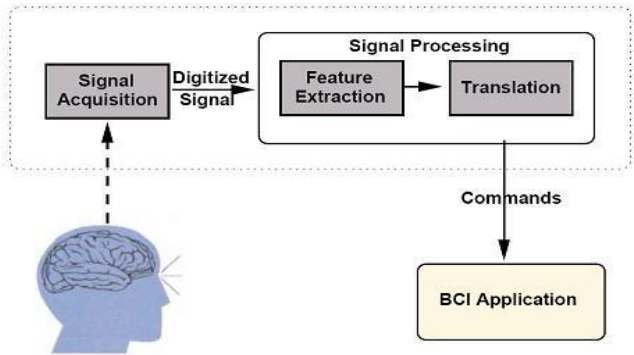

Fig. 1. Feature extraction and translation into machine commands

The author of [13] used features produced by Motor Imagery (MI) to control a robot arm. Features such as the band power in specific frequency bands (alpha: 8-12Hz and beta: 13-30Hz) were mapped into right and left limb movements. In addition, they used similar features with MI, which are the Event Related Desynchronization and Synchronization (ERD/ERS) comparing the signal's energy in specific frequency bands with respect to the mentally relaxed state. It was shown in [14] that the combination of ERD/ERS and Movement-Related Cortical Potentials (MRCP) improves EEG classification as this offers an independent and complimentary information.

In [15], a hybrid BCI control strategy is presented. The authors expanded the control functions of a P300 potential based BCI for virtual devices and MI related sensorimotor rhythms to navigate in a virtual environment. Imagined left/right hand movements were translated into movement commands in a virtual apartment and an extremely high testing accuracy results were reached.

A three-class BCI system was presented in [16] for the translation of imagined left/right hands and foot movements into commands that operates a wheelchair. This work uses many spatial patterns of ERD on mu rhythms along the sensory-motor cortex and the resulting classification accuracy for online and offline tests was 79.48% and 85.00%, respectively. The authors of [17] proposed an EEG-based BCI system that controls hand prosthesis of paralyzed people by movement thoughts of left and right hands. They reported an accuracy of about 90%.

A single trial right/left hand movement classification is reported in [18]. The authors analyzed both executed and imagined hand movement EEG signals and created a feature vector consisting of the ERD/ERS patterns of the mu and beta rhythms and the coefficients of the autoregressive model. Artificial Neural Networks (ANNs) is applied to two kinds of testing datasets and an average recognition rate of 93% is achieved.

The strength of BCI applications depends lies in the way we translate the neural patterns extracted from EEG into machine commands. The improvement of the interpretation of these EEG signals has become the goal of many researchers; hence, our research work explores the possibility of multi-trial EEG classification between left and right hand movements in an offline manner, which will enormously smooth the path leading to online classification and reading of any executed movements, leading us to what we can technically call "Reading Minds".

## III. THE PHYSIONET EEG DATA

### A. Description of the Dataset

The EEG dataset used in this research was created and contributed to PhysioNet [19] by the developers of the BCI2000 [20] instrumentation system. The dataset is publically available at http://www.physionet.org/pn4/eegmmidb/.

The dataset consists of more than 1500 EEG records, with different durations (one or two minutes per record), obtained from 109 healthy subjects. Subjects were asked to perform different motor/imagery tasks while EEG signals were recorded from 64 electrodes along the surface of the scalp. Each subject performed 14 experimental runs:

- A one-minute baseline runs (with eyes open)
- A one-minute baseline runs (with eyes closed)
- Three two-minute runs of each of the four following tasks:
  - The left or right side of the screen shows a target. The subject keeps opening and closing the corresponding fist until the target disappears. Then he relaxes.
  - The left or right side of the screen shows a target. The subject imagines opening and closing the corresponding fist until the target disappears. Then he relaxes.
  - The top or bottom of the screen. A target appears on either. The subject keeps opening and closing either both fists (in case of a top-target) or both feet (in case of a bottom-target) until the target disappears. Then he relaxes.
  - The top or bottom of the screen A target appears on either. The subject imagines opening and closing either both fists (in case of a top-target) or both feet (in case of a bottom-target) until the target disappears. Then he relaxes.

The 64-channels EEG signals were recorded according to the international 10-20 system (excluding some electrodes) as seen in Fig. 2.

*B. The Subset used in the Current Work*

From this dataset, we selected the three (two-minute) runs of the first task described above (opening and closing the left/right fist based on a target that appears on left or right side of the screen). These runs include EEG data for executed hand movements.

We created an EEG data subset corresponding to the first six subjects (S001, S002, S003, S004, S005, and S006) including three runs of executed movement specifically per subject for a total of 18 two-minute records.

IV. AUTOMATED ANALYSIS OF EEG SIGNALS FOR FEATURE EXTRACTION

*A. Channel Selection*

According to [6], many of the EEG channels appeared to represent redundant information. It is shown in [21, 22] that the neural activity that is correlated to the executed left and right hand movements is almost exclusively contained within the channels C3, C4, and CZ of the EEG channels of Fig. 2. This means that there is no need to analyze all 64 channels of data.

On the other hand, only eight electrode locations are commonly used for MRCP analysis covering the regions between frontal and central sites (FC3, FCZ, FC4, C3, C1, CZ, C2, and C4) [14]. These channels were used for the Independent Component Analysis (ICA) discussed later in the current section (Fig. 3).

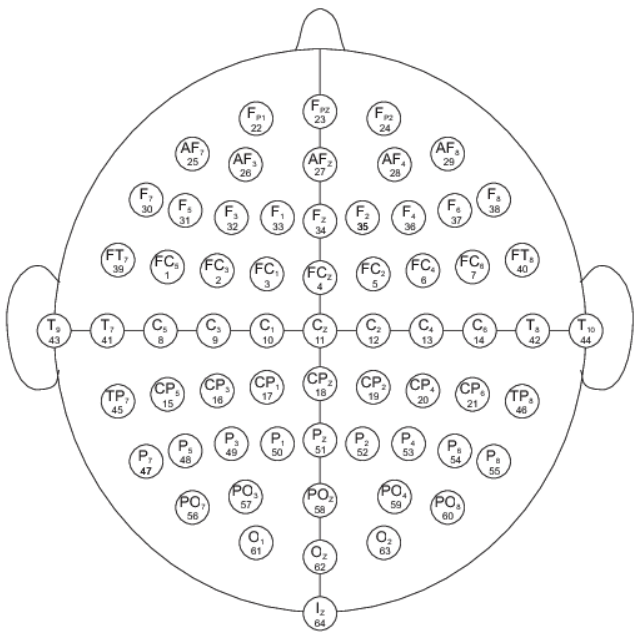

Fig. 2. Electrodes of the International 10-20 system for EEG

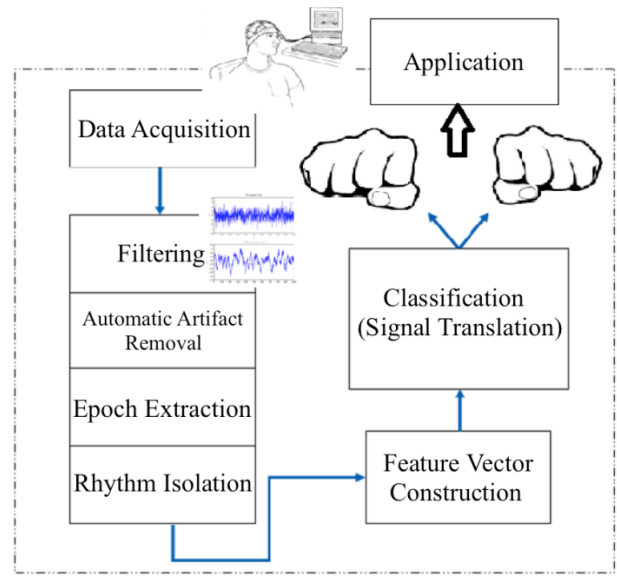

Fig. 3. Schematic diagram for the proposed system.

*B. Filtering*

Because EEG signals are known to be noisy and non-stationary, filtering the data is an important step to get rid of unnecessary information from the raw signals. EEGLAB [23], which is an interactive MATLAB toolbox, was used to filter EEG signals.

A band pass filter from 0.5 Hz to 90 Hz was applied to remove the DC (direct current) shifts and to minimize the presence of filtering artifacts at epoch boundaries. A Notch filter was also applied to remove the 50 Hz line noise.

*C. Automatic Artifact Removal (AAR)*

The EEG data of significance is usually mixed with huge amounts of useless data produced by physiological artifacts that masks the EEG signals [24]. These artifacts include eye and muscle movements and they constitute a challenge in the field of BCI research. AAR automatically removes artifacts from EEG data based on blind source separation and other various algorithms.

The AAR toolbox [25] was implemented as an EEGLAB plug-in in MATLAB and was used to process our EEG data subset on two stages: Electrooculography (EOG) removal using the Blind Source Separation (BSS) algorithm then Electromyography (EMG) Removal using the same algorithm [26].

*D. Epoch Extraction (Splitting)*

After the AAR process, the continuous EEG data were epoched by extracting data epochs that are time locked to specific event types.

When no sensory inputs or motor outputs are being processed, the mu (8–12 Hz) and beta (13–30 Hz) rhythms are said to be synchronized [4, 27]. These rhythms are electrophysiological features that are associated with the brain's normal motor output channels [4, 27]. While preparing for a movement or executing a movement, a desynchronization of the mu and beta rhythms occurs which is referred to as ERD and it can be extracted 1-2 seconds before onset of movement (as depicted in Fig. 4). Later, these rhythms synchronize again within 1-2 seconds after movement, and this is referred to as ERS.

On the other hand, delta rhythms can be extracted from the motor cortex, within the pre-movement stage, and this is referred to MRCP. The slow (less than 3 Hz) MRCP is associated with an event-related negativity that occurs 1-2 seconds before the onset of movement [28, 29].

In our experiments, we extracted time-locking events with type = 3 (left hand) or type = 4 (right hand) with different epoch limits and types of analysis:

- ERD analysis: epoch limits from -2 to 0 seconds.
- ERS analysis: epoch limits from 4.1 to 5.1 seconds.
- MRCP analysis: epoch limits from -2 to 0 seconds.

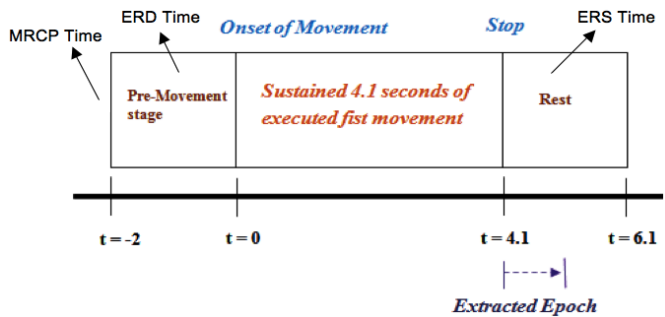

Fig. 4. Epoch Extraction (ERS/ERD and MRCP)

### E. Independent Component Analysis (ICA)

After the AAR process, ICA was used to parse the underlying electrocortical sources from EEG signals that are affected by artifacts [30, 31]. Data decomposition using ICA changes the basis linearly from data that are collected at single scalp channels to a spatially transformed virtual channel basis. Each row of the EEG data in the original scalp channel data represents the time course of accumulated differences between source projections to a single data channel and one or more reference channels [32].

EEGLAB was used to run ICA on the described epoched datasets (left and right ERD, ERS, and MRCP) for the channels FC3, FCZ, FC4, C3, C1, CZ, C2, and C4.

### F. Rhythm Isolation

A short IIR band pass filter from 8 to 30 Hz was applied on the ERD/ERS epoched datasets of the experiment for the purpose of isolating mu/beta rhythms. Another short IIR lowpass filter of 3 Hz was applied on MRCP epoched datasets for isolating delta rhythms. The result of this was 6 files for each run: ERD/ERS and MRCP for both left and right hand movements for each subject.

## V. PRACTICAL IMPLEMENTATION AND RESULTS

### A. Feature Vectors Construction and Numerical Representation

After the EEG datasets were analyzed as described in the previous section, the activation vectors were calculated for each of the resulted epochs' datasets as the multiplication of the ICA weights and ICA sphere for each dataset subtracting the mean of the raw data from the multiplication results.

Then, the mean, power, and energy of the activations were calculated to construct the feature vectors. For each subject's single run, 6 feature vectors were extracted as <Power (8 features), Mean (8 features), Energy (8 features), Type (1 feature: ERS/ERD/MRCP), Side (1 target: Left/Right)> resulting in a 108×26 feature matrix.

The constructed features were represented in a numerical format that is suitable for use with machine learning algorithms [33, 34]. Every column in the features matrices was normalized between 0.1 and 0.9 such that the datasets could be inputted to the learning algorithms described in the next subsection.

### B. Machine Learning Algorithms

In this work, Neural Networks (NNs) and Support Vector Machines (SVMs) algorithms were optimized for the purpose of classifying EEG signals into right and left hand movements. A detailed description of these learning algorithms can be found in [35] and [36].

The MATLAB neural networks toolbox was used for all NN experiments. The number of input features (25 features) determined the number of input nodes for NN and the number of different target functions (1 output: left or right) determined the number of output nodes. Training was handled with the aid of the back-propagation learning algorithm [37].

All SVM experiments were carried out using the "MySVM" software [38]. SVM can be performed with different kernels and most of them were reported to provide similar results for similar applications [6]. So, the Anova-Kernel SVM was used in this work.

### C. Optimisation and Results

In all experiments, 80% samples were randomly selected and used for training and the remaining 20% for testing. This was repeated 10 times, and in each time the datasets were randomly mixed.

For each experiment, the number of hidden nodes for NN varied from 1 to 20. In SVM, each of the degree and gamma parameters varied from 1 to 10. The mean of the accuracy was calculated for each ten training-testing pairs.

The features that were used as inputs to NN and SVM are symbolized as follows:

- P: the power.
- M: the mean.
- E: the energy.

- X: the sample type (ERS/ERD/MRCP).

The results of the experiment are summarized in the Table I.

TABLE I. RESULTS FOR EXPERIMENT B.

| Features | NN | | SVM | | |
|---|---|---|---|---|---|
| | Accuracy % | Hidden Layers | Accuracy % | Degree | Gamma |
| All | 88.9 | 3 | 85.3 | 1 | 5 |
| P, X | 80.4 | 15 | 88.2 | 3 | 4 |
| M, X | 68.5 | 11 | 91.2 | 3 | 10 |
| E, X | 82.1 | 11 | 94.1 | 4 | 5 |
| P, M, X | 79.8 | 3 | 80.6 | 8 | 3 |
| M, E, X | 82.7 | 9 | 82.4 | 5 | 4 |
| P, E, X | 89.8 | 4 | 97.1 | 4 | 4 |

It is clear from the testing results that SVM outperforms NN in most experiments. An SVM topology of degree = 4 and gamma = 4 provides an accuracy of 97.1% if tested with the power, energy and type inputs of the experiment. A NN of 10 hidden layers can provide an accuracy of 86.5% if all features are used. These results clearly show that the use of advanced feature extraction techniques provides good and clear properties that can be translated using machine learning into machine commands.

The next best SVM performance (94.1%) is achieved using the energy and type features. In general, there has been an increase in the classification performance with the use of more discriminative features, such as the total energy, compared to the power and mean inputs.

## VI. CONCLUSIONS AND FUTURE RESEARCH

This paper focuses on the classification of EEG signals for right and left fist movements based on a specific set of features. Very good results were obtained using NNs and SVMs showing that offline discrimination between right and left movement, for executed hand movements, is comparable to leading BCI research. Our methodology is not the best, but is somewhat a simplified efficient one that satisfies the needs for researchers in field of neuroscience.

In the near future, we aim to develop and implement our system in online applications, such as health systems and computer games. In addition, more datasets has to be analyzed for a better knowledgeable extraction and more accurate decision rules.

ACKNOWLEDGMENT

The authors would like to acknowledge the financial support received from Applied Science University that helped in accomplishing the work of this article.

References

[1] J. P. Donoghue, "Connecting cortex to machines: recent advances in brain interfaces," Nature Neuroscience Supplement, vol. 5, pp. 1085–1088, 2002.

[2] S. Levine, J. Huggins, S. BeMent, R. Kushwaha, L. Schuh, E. Passaro, M. Rohde, and D. Ross, "Identification of electrocorticogram patterns as the basis for a direct brain interface," Journal of Clinical Neurophysiology, vol. 16, pp. 439-447, 1999.

[3] A. Vallabhaneni, T. Wang, and B. He, "Brain—Computer Interface," in Neural Engineering, B. He, Ed.: Springer US, 2005, pp. 85-121.

[4] J. Wolpaw, N. Birbaumer, D. McFarland, G. Pfurtscheller, and T. Vaughan, "Brain-computer interfaces for communication and control," Clinical Neurophysiology, vol. 113, pp. 767-791, 2002.

[5] E. Niedermeyer and F. H. L. da Silva, Electroencephalography: Basic Principles, Clinical Applications, and Related Fields: Lippincott Williams & Wilkins, 2005.

[6] J. Sleight, P. Pillai, and S. Mohan, "Classification of Executed and Imagined Motor Movement EEG Signals," Ann Arbor: University of Michigan, 2009, pp. 1-10.

[7] B. Graimann, G. Pfurtscheller, and B. Allison, "Brain-Computer Interfaces: A Gentle Introduction," in Brain-Computer Interfaces: Springer Berlin Heidelberg, 2010, pp. 1-27.

[8] A. E. Selim, M. A. Wahed, and Y. M. Kadah, "Machine Learning Methodologies in Brain-Computer Interface Systems," in Biomedical Engineering Conference, 2008, CIBEC 2008. Cairo, 2008, pp. 1-5.

[9] E. Grabianowski, "How Brain-computer Interfaces Work," http://computer.howstuffworks.com/brain-computer-interface.htm, 2007.

[10] M. Smith, G. Salvendy, K. R. Müller, M. Krauledat, G. Dornhege, G. Curio, and B. Blankertz, "Machine Learning and Applications for Brain-Computer Interfacing," in Human Interface and the Management of Information. Methods, Techniques and Tools in Information Design. vol. 4557: Springer Berlin Heidelberg, 2007, pp. 705-714.

[11] J. J. Vidal, "Toward Direct Brain-Computer Communication," Annual Review of Biophysics and Bioengineering, vol. 2, pp. 157-180, 1973.

[12] G. Pfurtscheller, C. Neuper, D. Flotzinger, and M. Pregenzer, "EEG-based discrimination between imagination of right and left hand movement," Electroencephalography and Clinical Neurophysiology, vol. 103, pp. 642-651, 1997.

[13] F. Sepulveda, "Brain-actuated Control of Robot Navigation," in Advances in Robot Navigation, A. Barrera, Ed.: InTech, 2011.

[14] A.-K. Mohamed, "Towards improved EEG interpretation in a sensorimotor BCI for the control of a prosthetic or orthotic hand," in Faculty of Engineering. Master of Science in Engineering, Johannesburg: Universityof Witwatersrand, 2011, p. 144.

[15] Y. Su, Y. Qi, J.-x. Luo, B. Wu, F. Yang, Y. Li, Y.-t. Zhuang, X.-x. Zheng, and W.-d. Chen, "A hybrid brain-computer interface control strategy in a virtual environment," Journal of Zhejiang University SCIENCE C, vol. 12, pp. 351-361, 2011.

[16] Y. Wang, B. Hong, X. Gao, and S. Gao, "Implementation of a Brain-Computer Interface Based on Three States of Motor Imagery," in 29th Annual International Conference of the IEEE Engineering in Medicine and Biology Society, EMBS2007, 2007, pp. 5059-5062.

[17] C. Guger, W. Harkam, C. Hertnaes, and G. Pfurtscheller, "Prosthetic Control by an EEG-based Brain- Computer Interface (BCI)," in AAATE 5th European Conference for the Advancement of Assistive Technology, Düsseldorf, Germany, 1999.

[18] J. A. Kim, D. U. Hwang, S. Y. Cho, and S. K. Han, "Single trial discrimination between right and left hand movement with EEG signal," in Proceedings of the 25th Annual International Conference of the IEEE Engineering in Medicine and Biology Society, 2003., Cancun, Mexico, 2003, pp. 3321-3324 Vol.4.

[19] A. L. Goldberger, L. A. N. Amaral, L. Glass, J. M. Hausdorff, P. C. Ivanov, R. G. Mark, J. E. Mietus, G. B. Moody, C.-K. Peng, and H. E. Stanley, "PhysioBank, PhysioToolkit, and PhysioNet: Components of a New Research Resource for Complex Physiologic Signals," Circulation, vol. 101, pp. e215-e220, 2000.

[20] G. Schalk, D. J. McFarland, T. Hinterberger, N. Birbaumer, and J. R. Wolpaw, "BCI2000: a general-purpose brain-computer interface (BCI) system," IEEE Transactions on Biomedical Engineering, vol. 51, pp. 1034-1043, 2004.

[21] L. Deecke, H. Weinberg, and P. Brickett, "Magnetic fields of the human brain accompanying voluntary movements: Bereitschaftsmagnetfeld," Experimental Brain Research, vol. 48, pp. 144-148, 1982.

[22] C. Neuper and G. Pfurtscheller, "Evidence for distinct beta resonance frequencies in human EEG related to specific sensorimotor cortical areas," Clinical Neurophysiology, vol. 112, pp. 2084-2097, 2001.

[23] A. Delorme and S. Makeig, "EEGLAB: an open source toolbox for analysis of single-trial EEG dynamics," Journal of Neuroscience Methods, vol. 134, pp. 9-21, 2004.

[24] G. Bartels, S. Li-Chen, and L. Bao-Liang, "Automatic artifact removal from EEG - a mixed approach based on double blind source separation and support vector machine," in 2010 Annual International Conference of the IEEE Engineering in Medicine and Biology Society (EMBC), 2010, pp. 5383-5386.

[25] G. Gómez-Herrero, "Automatic Artifact Removal (AAR) toolbox for MATLAB," in Transform methods for Electroencephalography (EEG): http://kasku.org/projects/eeg/aar.htm, 2008.

[26] C. Joyce, I. Gorodnitsky, and M. Kutas, "Automatic removal of eye movement and blink artifacts from EEG data using blind component separation," Psychophysiology, vol. 41, pp. 313-325, 2004.

[27] A. Bashashati, M. Fatourechi, R. Ward, and G. Birch, "A survey of signal processing algorithms in brain-computer interfaces based on electrical brain signals," Journal of Neural Engineering, vol. 4, pp. R32-57, 2007.

[28] A. Vuckovic and F. Sepulveda, "Delta band contribution in cue based single trial classification of real and imaginary wrist movement," Medical and Biological Engineering and Computing, vol. 46, pp. 529 – 539, 2008.

[29] Y. Gu, K. Dremstrup, and D. Farina, "Single-trial discrimination of type and speed of wrist movements from EEG recordings," Clinical Neurophysiology, vol. 20, pp. 1596–1600, 2009.

[30] J. T. Gwin and D. Ferris, "High-density EEG and independent component analysis mixture models distinguish knee contractions from ankle contractions," in 2011 Annual International Conference of the IEEE Engineering in Medicine and Biology Society, EMBC, boston, USA, 2011, pp. 4195-4198.

[31] S. Makeig, A. J. Bell, T. P. Jung, and T. J. Sejnowski, "Independent component analysis of electroencephalographic data," Advances in Neural Information Processing Systems, vol. 8, pp. 145-151, 1996.

[32] A. Delorme and S. Makeig, "Single subject data processing tutorial: Decomposing Data Using ICA," in The EEGLAB Tutorial: http://sccn.ucsd.edu/wiki/EEGLAB, 2013.

[33] M. Al-Omari, R. Qahwaji, T. Colak, and S. Ipson, "Machine learning-based investigation of the associations between cmes and filaments," Solar Physics, vol. 262, pp. 511-539, 2010.

[34] R. Qahwaji, T. Colak, M. Al-Omari, and S. Ipson, "Automated machine learning based prediction of CMEs based on flare associations," Sol. Phys, vol. 248, 2007.

[35] R. Qahwaji and T. Colak, "Automatic Short-Term Solar Flare Prediction Using Machine Learning and Sunspot Associations," Solar Phys., vol. 241, pp. 195-211, 2007.

[36] R. Qahwaji, M. Al-Omari, T. Colak, and S. Ipson, "Using the Real, Gentle and Modest AdaBoost Learning Algorithms to Investigate the Computerised Associations between Coronal Mass Ejections and Filaments," in Mosharaka International Conference on Communications, Computers and Applications (MIC-CCA 2008), Mosharaka for Researches and Studies, Amman, Jordan, 2008, pp. 37-42.

[37] S. E. Fahlmann and C. Lebiere, "The cascade-correlation learning architecture," in Advances in Neural Information Processing Systems 2 (NIPS-2) Denver, Colorado, 1989.

[38] S. Rüping, "mySVM-Manual ": University of Dortmund, Lehrstuhl Informatik 8, 2000.